%% file: main.tex
\pdfoutput=1

\documentclass[11pt]{article}

\usepackage{natbib}

\usepackage[review]{acl}

\usepackage{times}
\usepackage{pgfplots}
\usepackage{fancyhdr} 
\usepackage{latexsym}
\usepackage{amsmath} 
\usepackage{graphicx} 
\usepackage{float} 
\usepackage{pgf-pie} 

\usepackage{xcolor} 
\definecolor{gray1}{RGB}{200,200,200} 
\definecolor{gray2}{RGB}{180,180,180} 
\definecolor{gray3}{RGB}{160,160,160} 
\definecolor{gray4}{RGB}{140,140,140} 
\definecolor{gray5}{RGB}{120,120,120} 
\definecolor{gray6}{RGB}{100,100,100} 
\definecolor{gray7}{RGB}{80,80,80}    

\usepackage[normalem]{ulem}
\usepackage[T1]{fontenc}

\usepackage[utf8]{inputenc}

\usepackage{microtype}

\usepackage{caption}
\usepackage{subcaption}
\usepackage{inconsolata}
\usepackage{hyperref}

\usepackage{enumitem}
\setlist[enumerate]{itemsep=2pt, parsep=0pt}  

\newcommand{\thickhline}{\noalign{\hrule height 1pt}}
\title{DeepScore: A Comprehensive Approach to Measuring Quality in AI-Generated Clinical Documentation}

\author{%
Jon Oleson
\\ 
\textit{DeepScribe Inc.}\\
\textit{San Francisco, California, USA}\\
\textit{jon.oleson@deepscribe.ai}\\
}

\begin{document}
\makeatletter
\newif\ifacl@finalcopy
\acl@finalcopytrue
\makeatother

\maketitle

\begin{abstract}
Medical practitioners are rapidly adopting generative AI solutions for clinical documentation, leading to significant time savings and reduced stress. However, evaluating the quality of AI-generated documentation is a complex and ongoing challenge. This paper presents an overview of DeepScribe's methodologies for assessing and managing note quality, focusing on various metrics and the composite "DeepScore", an overall index of quality and accuracy. These methodologies aim to enhance the quality of patient care documentation through accountability and continuous improvement.
\end{abstract}

\vspace{-10pt}
\input{section1-intro}

\section{Methods}

\input{section2.1-statrates}

\input{section2.2-recallprecision}

\input{section2.3-useracceptance}

\input{section2.4-transcriptionqc}

\input{section2.5-deepscore}

\section{Limitations}
\input{section3-limitations}

\section{Future Work}
\input{section4-futurework}

\section{Conclusion}
\input{section5-conclusion}

\bibliographystyle{plainnat}
\bibliography{main}

\end{document}

%% file: section1-intro.tex
\section{Introduction}
In the evolving healthcare landscape, generative AI for clinical documentation is seeing rapid adoption, offering significant time savings and reducing practitioner burnout. Despite these immense benefits, evaluating the quality of AI-generated documentation remains complex. Current accuracy metrics, such as Word Error Rate, F1 Scores, and Precision/Recall \citep{schloss2020automatedsoapnoteclassifying}, as well as recent innovations like BERTScore \citep{li2024improvingclinicalnotegeneration}, do not fully address the complexities and potential risks in clinical settings. This paper presents DeepScribe’s methodologies for assessing the quality of AI-generated medical documentation, using comprehensive metrics to capture both technical completeness and accuracy, as well as less tangible concepts like user acceptance. The Major Defect-Free Rate (MDFR) and Critical Defect-Free Rate (CDFR) assess significant and critical error frequencies. The Captured Entity Rate (CER) and Accurate Entity Rate (AER) measure the relevance and precision of captured medical information. The Minimally-Edited Note Rate (MNR) reflects user acceptance through content-editing behaviors. The Medical Word Hit Rate (MWHR) evaluates the accuracy of transcribed medical terms. These metrics are used to derive the composite "DeepScore", representing the overall quality of autonomous transcription and scribing, guiding continuous improvement in patient care documentation.

\begin{table}[h]
\begin{center}
\small
\setlength{\tabcolsep}{15pt}  
\renewcommand{\arraystretch}{1.5}  
\begin{tabular}{r c c l}
\thickhline
Metric & Value & Category & Source \\
\hline
MDFR & 95.9\% & Stat Rates & Stat Rates Run, March 2024 \\
CDFR & 100.0\% & Stat Rates & Stat Rates Run, March 2024 \\
CER & 90.2\% & Recall/Precision & Stat Rates Run, March 2024 \\
AER & 96.2\% & Recall/Precision & Stat Rates Run, March 2024 \\
MNR & 95.0\% & User Acceptance & Production Note Edit Rates, 1/15/2024 - 6/12/2024 \\
MWHR & 95.3\% & Transcription QC & Transcription QC Sample, May 2024 \\
\hline
\textbf{DeepScore} & \textbf{95.4\%} & Composite & Note Quality Report, May 2024 \\
\thickhline
\end{tabular}
\caption {Note Quality Scorecard, June 7 2024}
\vspace{-15pt}
\label{table:1}
\end{center}
\end{table}

\newpage

%% file: section2.1-statrates.tex
\subsection{Statistical Rates}

At DeepScribe, we use Statistical Rates (Stat Rates) to track progress in overall model performance and to pinpoint areas of model regression, enabling our product and engineering teams to focus development effort where it can have the most impact.

\subsubsection{Results}

\begin{table}[h]
\begin{center}
\small

\begin{tabular}{c c}

\thickhline
Metric & Value \\
\hline
Major Defect-Free Rate (MDFR) & 95.9\%\\
Critical Defect-Free Rate (CDFR) & 100.0\%\\
\thickhline
\end{tabular}

\caption {Source: Monthly Stat Rates, 100-Note Test Set, March 2024.}

\label{table:1}

\end{center}
\end{table}

\subsubsection{Overview}
Stat Rates is a DeepScribe system designed to evaluate the quality of medical notes by scrutinizing note content for completeness and accuracy. At its core, this method involves a detailed comparison of AI-scribed notes against a 'rubric'—a note scribed by a human expert and vetted by peers— derived from the original audio of a clinical encounter. The AI-scribed note is assessed for defects which are classified by type and severity, ranging from minor inaccuracies to critical errors that could impact patient safety.

This granular approach enables precise feedback that guides both the immediate corrections needed and broader developmental improvements in DeepScribe AI algorithms, ensuring that the system evolves to meet the high standards of medical documentation required for clinical efficacy and patient care.

\subsubsection{Key Concepts}

\begin{itemize}
    \item \textbf{Test Encounter:} The foundation of Stat Rates is the evaluation of ’test encounters’—real but completely de-identified interactions between clinicians and patients, recorded and transcribed.
    \item \textbf{Test Note:} The ’test note’ generated by DSX, our product, from the test encounter transcript represents the practical output of our system, segmented into ’snippets’ of information for detailed assessment.
    \item \textbf{Test Set:} The full collection of test notes that we evaluate when running Stat Rates is called the 'Test Set'. We have constructed the Test Set to be a representative sample of our production notes, but periodically must update its composition to reflect the evolving mix of notes we scribe in production.
    \item \textbf{Rubric:} Each test note is compared against a 'rubric', or a note that has been manually written from the same encounter's transcript by an expert human scribe, representing the best available human standard.
    \item \textbf{Entity:} A rubric is parsed into ’entities’, key pieces of medically relevant content that the test note must reflect accurately. 
    \item \textbf{Snippet:} A piece of medically relevant content in the test note, identified by an auditor. Each snippet should correspond to an entity in the rubric.
\end{itemize}

\newpage
\vspace{5pt}
Example rubric, for illustration purposes only:
\begin{figure}[H]
    \centering
    \includegraphics[width=\linewidth, trim={2mm 2mm 0 0}, clip]{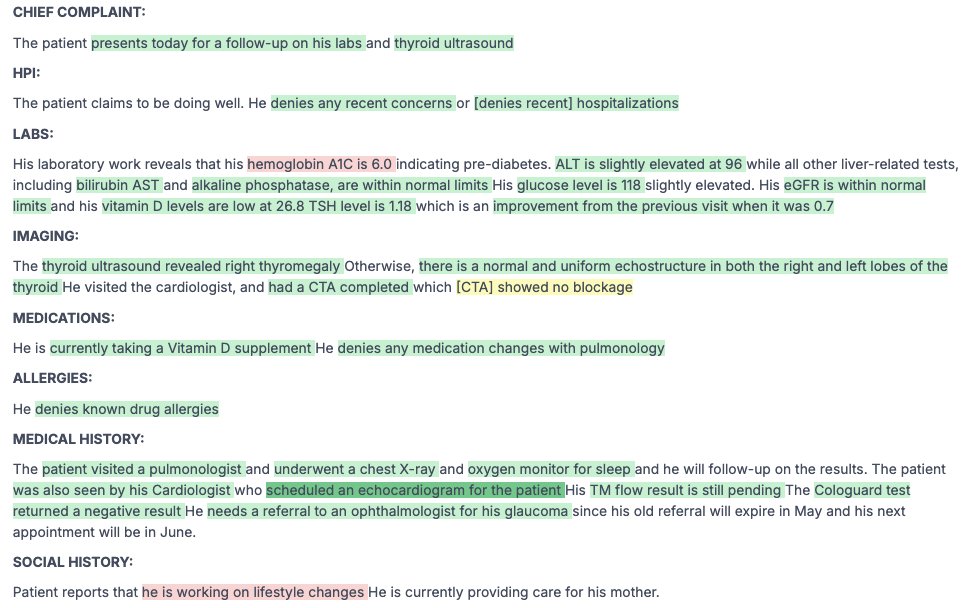}
    \caption{A rubric with highlighted entities.}
    \label{fig:rubric_example}
\end{figure}

\vspace{10pt}  

\noindent\textbf{Defects and Severity.} Discrepancies between the test note and the rubric are classified as defects. These are further evaluated and assigned a severity score according to the following \textbf{Defect Severity Scale}, with a priority on patient safety:

\vspace{5pt}  

\begin{enumerate}[topsep=0pt, partopsep=0pt]
  \item Low impact: The defect is so immaterial that it is unlikely to be noticed.
  \item Mild impact: The defect could potentially lead to misunderstandings or miscommunication but is still just annoying rather than actively harmful.
  \item Moderate impact: The defect poses a risk of moderate harm if not corrected, for example omitting some detail about a symptom relevant to the chief complaint.
  \item Major impact: The defect poses a significant risk of causing incorrect treatment or diagnosis, demanding immediate correction.
  \item Critical impact: The defect could lead to serious adverse patient outcomes without correction, such as incorrect surgery or medication.
\end{enumerate}

\vspace{5pt}  

\noindent\textbf{Auditing Methodology:}
To run Stat Rates, we distribute test notes amongst a team of auditors. To audit a given test note, an auditor scrutinizes each snippet, checking it for accuracy against the corresponding rubric entity. This approach ensures that every piece of clinical information is accounted for and correctly represented.

\newpage
\vspace{5pt}
Test note alongside connected rubric, for illustration purposes only:
\begin{figure}[H]
    \centering
    \includegraphics[width=\linewidth]{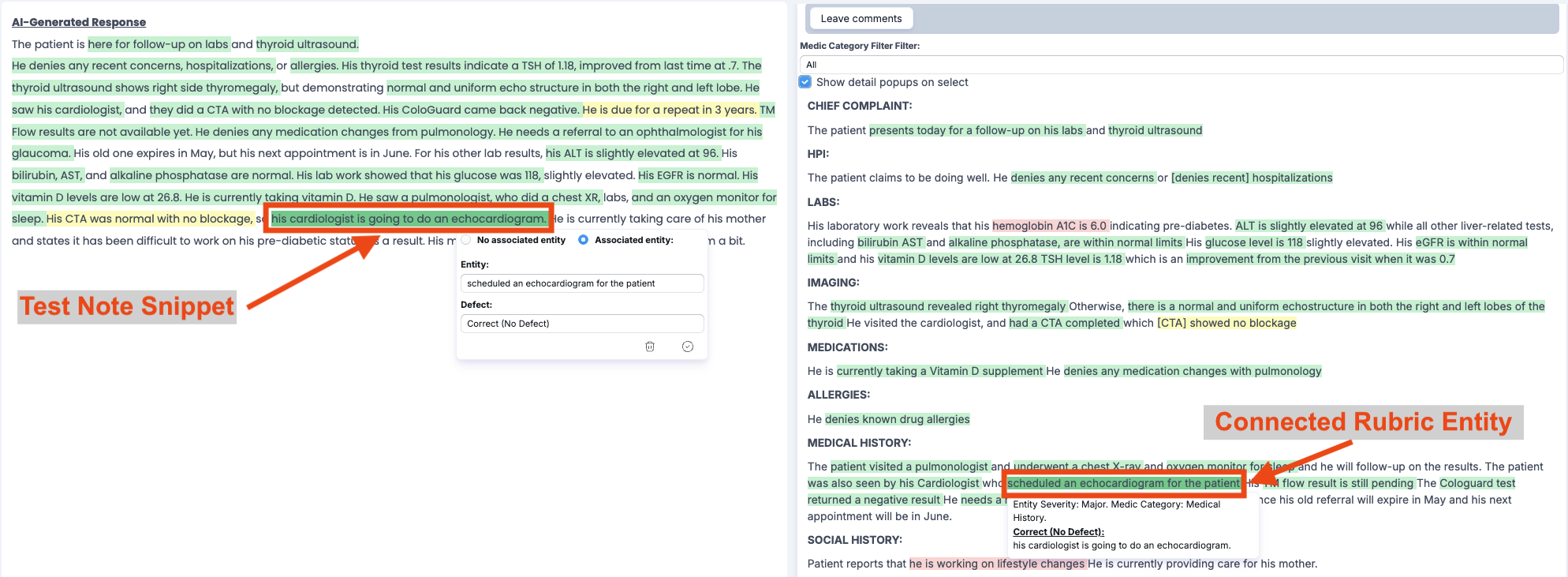}
    \caption{A test note (left) with snippets connected to rubric entities (right).}
    \label{fig:audit_example}
\end{figure}

\subsubsection{Metric Definitions}
\begin{itemize}
    \item \textbf{Major Defect-Free Rate (MDFR):} The percentage of entities in the test set that are free of defects with severity >= 4 (Major or Critical impact).
    \item \textbf{Critical Defect-Free Rate (CDFR):} The percentage of entities in the test set that are free of defects with severity = 5 (Critical impact).
\end{itemize}

%% file: section2.2-recallprecision.tex
\subsection{Recall to Precision Funnel}
The Recall to Precision Funnel is derived from Stat Rates results to provide a focused view on the relevance and accuracy of the note content. We first by run Stat Rates to produce the output dataset of entities from test notes. Each entity is tagged as correct or defective and categorized by medical relevance and defect severity. Then, for a given test encounter, we examine what proportion of the relevant information from the rubric was captured by the AI in the test note and, subsequently, what proportion of that included information was categorized and summarized correctly.

\subsubsection{Results}
\begin{table}[h]
\begin{center}
\small
\begin{tabular}{c c}
\thickhline
Metric & Value \\
\hline
Captured Entity Rate (CER) & 90.2\%\\
Accurate Entity Rate (AER) & 96.0\%\\
\thickhline
\end{tabular}
\caption {Source: Monthly Stat Rates, 100-Note Test Set, March 2024.}
\label{table:1}
\end{center}
\end{table}

\subsubsection{Metric Definitions}
\begin{itemize}
    \item \textbf{Missing Entity Rate (MER):} We calculate MER by identifying entities classified as Missing Information defects of Major or Critical severity. This metric reflects the proportion of key information that was omitted in the AI-generated note.
        \begin{itemize}
            \item \textbf{The Captured Entity Rate (CER)} (akin to "Recall") is derived from 1 - MER. It quantifies the percentage of medically relevant information captured in the AI generated note.
        \end{itemize}
    \item \textbf{Inaccurate Entity Rate (IER):} We calculate IER by identifying entities classified as Inaccurate with Major or Critical Severity. This metric reflects the proportion of key information that was captured in the AI-generated note, but incorrectly categorized or summarized.
        \begin{itemize}
            \item \textbf{The Accurate Entity Rate (AER)} (i.e., "Precision") is calculated from 1 - IER, representing the accuracy of the information captured.
        \end{itemize}   
\end{itemize}

\vspace{5pt}
Example CER calculation, for illustration purposes only:

\begin{figure}[H]
    \centering
    \includegraphics[width=\linewidth]{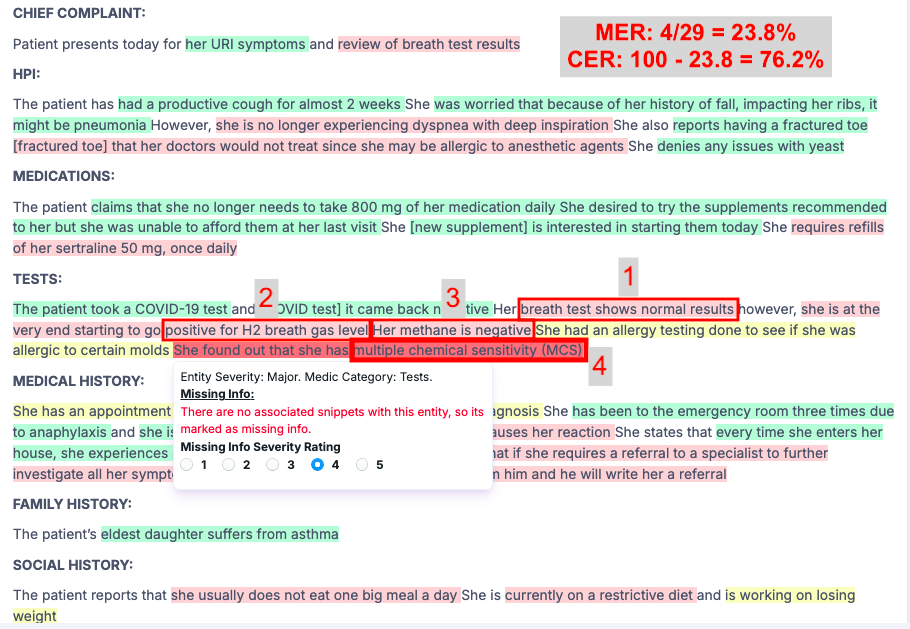}
    \caption{A demonstration of how CER is calculated on a single note.}
    \label{fig:recall_cer_example}
\end{figure}

\vspace{10pt}  

\vspace{5pt}
Example AER calculation, for illustration purposes only:

\begin{figure}[H]
    \centering
    \includegraphics[width=\linewidth]{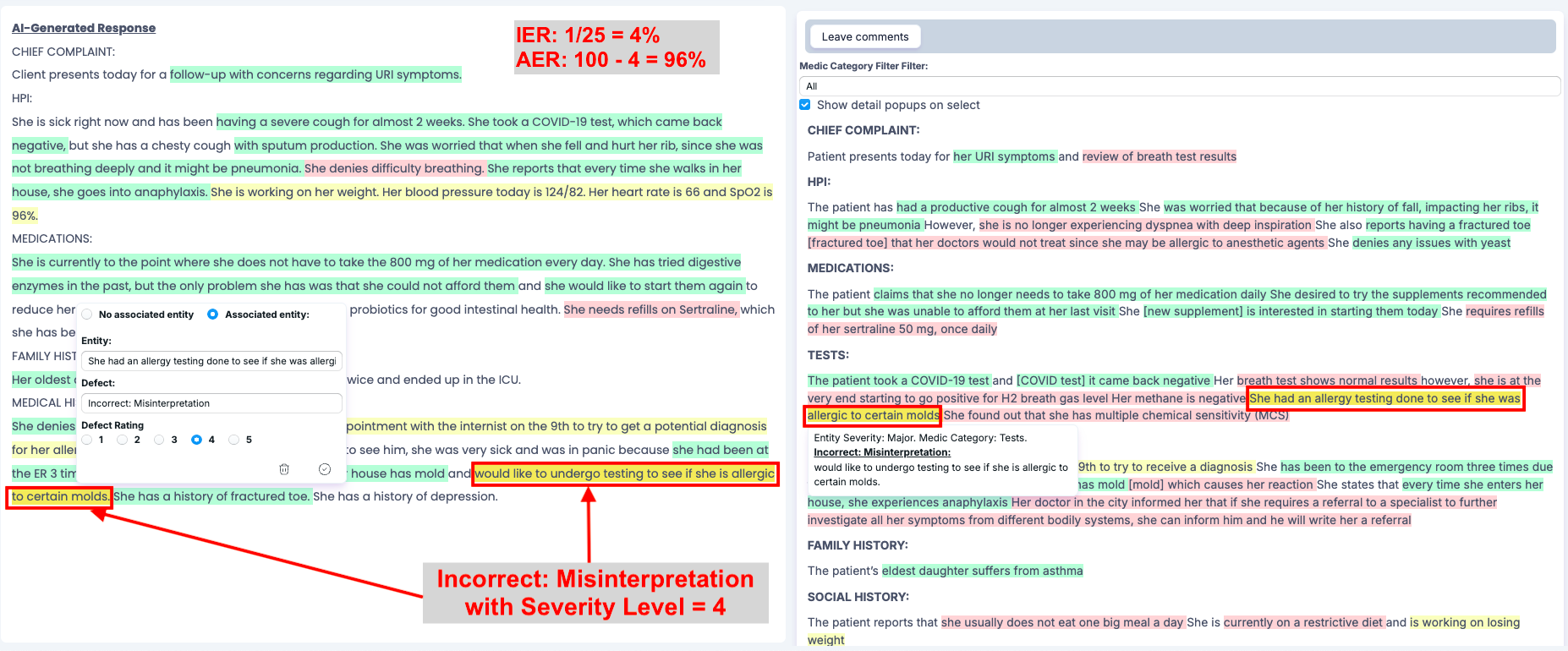}
    \caption{A demonstration of how AER is calculated on the same note, with reference to the rubric.}
    \label{fig:precision_aer_example}
\end{figure}

%% file: section2.3-useracceptance.tex
\subsection{User Acceptance}
User Acceptance of DeepScribe's transcribed notes is assessed by analyzing the editing behaviors of users, which serves as an indirect measure of the system's performance from the user's perspective.

\subsubsection{Results}
\begin{table}[H]
    \begin{center}
    \small
    \begin{tabular}{c c}
    \thickhline
    Metric & Value \\
    \hline
    Minimally-Edited Note Rate (MNR) & 95.0\%\\
    \thickhline
    \end{tabular}
    \caption {Source: Production Note Edit Metrics, 135.9k Notes, 1/15/2024 - 6/12/2024.}
    \label{table:1}
    \end{center}
\end{table}

\subsubsection{Key Concepts}

\begin{itemize}
    \item \textbf{Words Added:} Within the final version of the note, we track the percentage of words added by users, highlighting areas where additional information was deemed necessary.
    \item \textbf{Words Deleted:} This metric measures the percentage of words from the initial draft that users remove, which may indicate over-generation or irrelevance of content.
    \item \textbf{Words Substituted:} By examining the percentage of words from the initial draft that users substituted, we assess potential inaccuracies or preferences for different terminologies.
\end{itemize}

For our purposes we pay closest attention to substitutions, where content in the note is directly altered, rather than instances where new content is added or existing content is deleted. Clinicians may add content to notes that is completely beyond the context of the encounter, and may delete verifiably relevant content for unknown reasons, making metrics based on those behaviors less reliable.

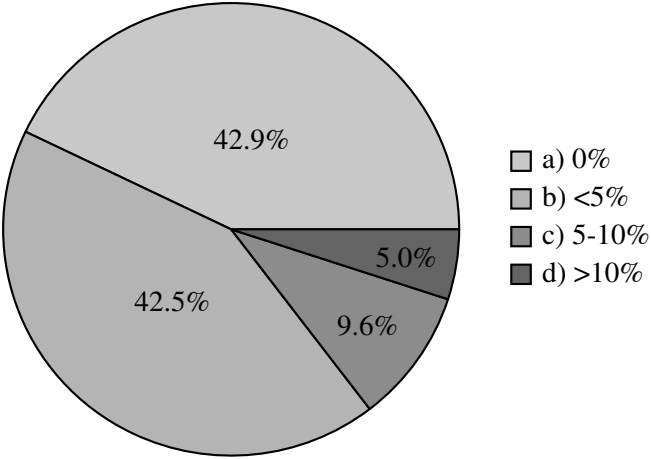
\begin{figure}[h]
\centering
\begin{tikzpicture}
\pie[text=legend, color={gray1, gray2, gray4, gray6}, radius=3]{
    42.9/a) 0\%, 42.5/b) <5\%, 9.6/c) 5-10\%, 5.0/d) >10\%
}
\end{tikzpicture}
\caption{Pct Notes in Words Substituted Rate Segment}
\label{fig:pie_chart}
\end{figure}

\subsubsection{Metric Definitions}
\begin{itemize}
    \item \textbf{Minimally-Edited Note Rate (MNR):} MNR is the percentage of notes where < 10\% of words in the note were substituted by the user.
\end{itemize}

%% file: section2.4-transcriptionqc.tex
\pdfoutput=1

\newpage
\subsection{Transcription Quality Control}
Transcription Quality Control measures the effectiveness of DeepScribe’s Automated Speech Recognition (ASR)  engine. The key metric we use in this process is the Medical Word Hit Rate (MWHR), which specifically measures the accuracy of medical terms transcribed. 

\subsubsection{Results}
\begin{table}[H]
    \begin{center}
    \small
    \begin{tabular}{c c}
    \thickhline
    Metric & Value \\
    \hline
    Medical Word Hit Rate (MWHR) & 95.3\%\\
    \thickhline
    \end{tabular}
    \caption {Source: Monthly Transcription QC, 10-Note Test Set, May 2024.}
    \label{table:1}
    \end{center}
\end{table}

\subsubsection{Metric Definitions}
\begin{itemize}
    \item \textbf{Medical Word Hit Rate (MWHR):} MWHR assesses the percentage of medical terms that are clearly audible in the source audio that are then correctly transcribed.
\end{itemize}

\noindent\textbf{Example:}
Here is a portion of an example transcript that includes 90 medical terms overall:

\begin{quote}
Speaker 1: So we're going to stop the \sout{client} Clonidine and do you get like hot flashes?

Speaker 2: No. Only when I'm sick, like it's the only time.

Speaker 1: And you're just on a program for immunosuppression, right? 

Speaker 2: Yeah. 

Speaker 1: For the \sout{invasive} Envarsus? 

Speaker 2: Yeah. That's fine.
\end{quote}

Of the 90 medical terms mentioned in this entire transcript, this portion shows 2 mis-transcriptions (“client” instead of “Clonodine” and “invasive” instead of “Envarsus”). Thus, in this example, MWHR = 88/90 = 97.8\%.

%% file: section2.5-deepscore.tex
\pdfoutput=1

\subsection{DeepScore}
\textbf{DeepScore} is a composite metric that combines all of the aforementioned metrics into a single score. To derive this, we calculate the average of the component metrics. 

\subsubsection{Results}
\begin{table}[h]
    \begin{center}
    \small
    \begin{tabular}{c c}
    \thickhline
    Metric & Value \\
    \hline
    \textbf{DeepScore} & \textbf{95.4\%}\\ 
    \thickhline
    \end{tabular}
    \caption {Source: Note Quality Report, May 2024.}
    \label{table:1}
    \end{center}
\end{table}

\vspace{-10pt}  

\subsubsection{Formula and Calculation}
\textbf {Formula:}

\begin{equation*}
\text{DeepScore} = [ \text{MDFR} 
                     + \text{CDFR}
                     + \text{CER}
                     + \text{AER}
                     + \text{MNR}
                     + \text{MWHR}
                    ] / 6
\end{equation*}

\noindent\textbf {Calculation:}

\begin{equation*}
\text{DeepScore} = [95.9 + 100.0 + 90.2 + 96.2 + 95.0 + 95.3] / 6
\end{equation*}
\[
= 95.4
\]

%% file: section3-limitations.tex
\pdfoutput=1

\subsection{Stat Rates and Recall → Precision Funnel Metric}
\begin{itemize}
    \item \textbf {Inherent Challenges:} These methodologies, while providing a granular analysis of note quality, may not fully encompass the complexity of human clinical judgment. They rely heavily on the predefined rubrics and defect severity scales which may not capture all nuances of medical dialogue or context.
    \item \textbf {Test Set Composition:} As DeepScribe evolves and launches support for new specialties, the composition of the existing Test Set may no longer be representative of the mix of note types in production. This limitation means that the metrics derived from the Test Set might not fully capture the performance and quality of the system across additional medical specialties and new types of clinical encounters.
    \item \textbf {Subjectivity in Severity Judgments:} The effectiveness of these metrics can be significantly influenced by human subjectivity in assigning defect severity levels. Variability in judgment among different reviewers assessing the same defects can lead to inconsistencies in the data, impacting the reliability and comparability of the outcomes.
    \item \textbf {Training and Calibration Challenges:} Consistency in defect severity scoring requires continuous training and calibration among evaluators. Without rigorous standardization, the subjective nature of defect classification can undermine the objective assessment of note quality.
\end{itemize}

\subsection{User Acceptance}
\begin{itemize}
    \item \textbf {Subjectivity and Variability:} This metric is influenced by individual user behavior and subjective judgment, which can vary widely among users. As such, high editing rates might reflect personal preference rather than objective errors, potentially skewing the data on system performance.
    \begin{itemize}
        \item For example, when adding words, users may frequently add context that is not part of the transcript.
        \item When deleting words, users may be dropping content that is relevant but that they don’t want cluttering the note for reasons of their own unique preference.
    \end{itemize}
\end{itemize}

\subsection{Transcription QC}
\begin{itemize}
    \item \textbf {Scope Limitation:} MWHR focuses exclusively on the accuracy of medical terms, which, while crucial, represents a fraction of the overall content of clinical notes. This narrow focus might overlook other types of errors that could also impact the usability and understanding of the notes.
    \item \textbf {Audio Quality Dependence:} The effectiveness of MWHR is contingent upon the quality of the audio input. Poor audio quality can lead to misinterpretations and incorrect transcriptions, significantly affecting the metric's reliability.
    \item \textbf {Small-Sample Volatility:} Due to the costly and time-intensive nature of calculating the metric, we relied on a fairly small sample of 10 notes to estimate MWHR. It is possible that a different or larger sample would result in a materially different estimate.
\end{itemize}

\subsection{DeepScore}
\begin{itemize}
    \item \textbf {Subjectivity of Unweighted Averaging:} The decision not to assign weights to the component metrics is subjective and may not correspond optimally to broader expert opinion.
    \item \textbf {Limited Validation:} DeepScore is a novel metric whose usefulness has not been validated by ongoing usage over multiple cycles.
\end{itemize}

%% file: section4-futurework.tex
\pdfoutput=1

Future work will focus on establishing this suite of quality metrics for distinct medical specialties. This will involve creating specialized test sets and evaluation criteria to address the unique documentation needs of each field, as well as developing specialty-specific formulations of the "DeepScore" composite.

%% file: section5-conclusion.tex
\pdfoutput=1

This paper presents a comprehensive framework for assessing the quality of AI-generated clinical documentation using DeepScribe’s multifaceted evaluation methodologies. By integrating various metrics—such as Major Defect-Free Rate (MDFR), Critical Defect-Free Rate (CDFR), Captured Entity Rate (CER), Accurate Entity Rate (AER), Minimally-Edited Note Rate (MNR), and Medical Word Hit Rate (MWHR)—we demonstrate a robust system for evaluating the performance of our AI medical
note-scribing system. The introduction of the composite "DeepScore" encapsulates these metrics into a single, interpretable measure of overall quality. With this system, we can pinpoint specific issues and maintain a high-level perspective on quality, facilitating targeted interventions while promoting continuous overall improvement of the DeepScribe platform.